\documentclass[10pt,twocolumn,letterpaper]{article}

\usepackage{cvpr}              

\usepackage{multirow}

\definecolor{cvprblue}{rgb}{0.21,0.49,0.74}
\usepackage[pagebackref,breaklinks,colorlinks,allcolors=cvprblue]{hyperref}

\def\paperID{11946} 
\def\confName{CVPR}
\def\confYear{2026}

\title{Inconsistency-aware Multimodal Schrödinger Bridge for Deepfake Localization}

\author{
Jiayu Xiong$^{1,2}$ \quad
Jing Wang$^{1,2}$\thanks{Corresponding authors.} \quad
Qi Zhang$^{3}$\footnotemark[1] \quad
Wanlong Wang$^{1,2}$ \quad
Jun Xue$^{4}$ \\
$^{1}$Department of Computer Science and Techonology, Huaqiao University \\
$^{2}$Xiamen Key Laboratory of Computer Vision and Pattern Recognition, Huaqiao University \\
$^{3}$Tongji University \quad
$^{4}$School of Cyber Science and Engineering, Wuhan University \\
{\tt\small \{yuinst, wwl58\}@stu.hqu.edu.cn, wroaring@hqu.edu.cn,} 
\\
{\tt\small zhangqi\_cs@tongji.edu.cn, junxue@whu.edu.cn}
}

\begin{document}
\maketitle

\begin{abstract}
Audio-visual deepfake localization demands interval-level outputs that serve as temporal evidence. Despite recent progress, symmetric fusion under single-sided or asynchronous forgeries propagates cross-modal noise, degrading high-precision localization. We present IaMSB, an inconsistency-aware multimodal Schrödinger Bridge (SB) that jointly estimates cross-modal consistency and performs interval-level localization. Unlike diffusion models, SB minimizes path-distribution discrepancy and yields consistency scores without explicit noise injection or denoising. With the Schrödinger Bridge (SB), IaMSB unifies consistency estimation, cross-modal information selection, and bridge-step scheduling in one framework. Specifically, a lightweight coarse bridge first proposes candidate intervals and estimates cross-modal consistency; these statistics select cross-modal witness signals and allocate bridge steps asymmetrically across modalities. A refinement bridge then performs step-tuned fusion and outputs refined, time-aligned intervals. IaMSB anticipates single-sided and asynchronous forgeries and, using bottlenecked cross-modal interaction with step allocation, suppresses noise transfer, avoids unnecessary iterations. Across benchmarks, IaMSB stabilizes strict-IoU boundary precision, raising AP@0.95 by $3\sim10\%$, and yields improved high-precision localization, particularly for single-sided forgeries.
\end{abstract}

\section{Introduction}
\label{sec:intro}
\begin{figure}
    \centering
    \includegraphics[width=1\linewidth]{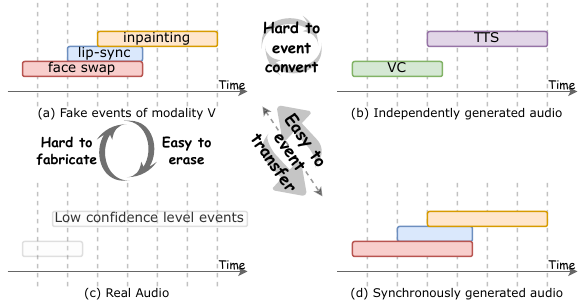}
    \caption{A bridge estimates the cross-modal deepfake event set's reachability. (a)(b): Mismatched events rarely reach; Subfigure (a)(c): single-sided cases are asymmetric, and (a)(d): consistent events reach more easily; arrival steps define the generation cost for suspicion scoring and budget allocation.}
    \vskip -1em
    \label{fig:csb}
\end{figure}
Rapid proliferation of synthetic media and falling tampering costs make deepfake localization essential for governance and forensics~\cite{mirsky2021deepfakes,verdoliva2020media,chesney2019deepfakes}. Interval-level predictions provide auditable, time-stamped evidence that surpasses video-level labels and directly supports moderation and legal adjudication~\cite{cai2023lavdf_cviu,zhang2023ummaformer}. Because audio and visual often diverge in confidence, existence of manipulation, onset, and duration, methods that explicitly resolve cross-modal asymmetry while delivering precise temporal boundaries are necessary~\cite{chugh2020mds,oorloff2024avff,yang2023avoiddf}.

Recent work has markedly advanced cross-modal localization via symmetric fusion modules~\cite{katamneni2024mmmsba,liu2024avtfd,zhang2023ummaformer,zhu2025regqav} (i.e., order-invariant, equal-depth fusion) with temporal refinement~\cite{cai2022lavdf_dicta,cai2023lavdf_cviu,wu2024cfprf,zhang2023ummaformer}. However, sample lengths and the occurrence of deepfake events are often unaligned, and temporal fusion for sequence modeling is computationally expensive. This yields three issues: \textbf{(i)} negative transfer from the clean (non-forged) modality, injecting noise and inducing false localization~\cite{Xu_2024_WACV,wei2024diagnosing,wu2024mlmm,yang2024facilitating}; and \textbf{(ii)} too many fusion layers waste compute on the clean modality, whereas too few hinder convergence of the forged modality~\cite{hu2025daf,Cao_2024_CVPR}; and \textbf{(iii)} given computational overhead of fusion~\cite{rahmath2025earlyexit,bajpai-hanawal-2024-ceebert,Wang_2024_CVPR}, an unavoidable trade-off constrains temporal resolution~\cite{zhu2025regqav}. Consequently, reliability remains constrained precisely in high-precision localization regimes.

Generative decoders allow explicit step control of inference budgets and can speed up convergence~\cite{wang2024fas,debortoli2021dsb}, which provides inspiration for asymmetric fusion (i.e., adaptive steps or equivalent layers). Framing event boundaries as multi-step denoising of noisy queries yields traceable generation trajectories and more stable training objectives~\cite{xu2024denoiseLoc,hwang2025diffgebd,chen2025diffdvc}. However, existing formulations still lack a principled scheduler for multimodal asynchrony and a calibrated, temporally local measure of cross-modal discrepancy to steer interaction and compute. These gaps point to a bridge-based decoder that couples consistency scoring with step-wise generation.

Motivated by this, we formulate audio–visual deepfake localization as event generation and introduce \textbf{IaMSB} (inconsistency-aware multimodal Schrödinger Bridge), a framework that first leverages diffusion Schrödinger Bridge (SB) for cross-modal consistency estimate, interaction, and deepfake event generation. SB cast as stochastic control, transports a source distribution to a target distribution without an explicit noise-adding/denoising cycle while directly quantifying distributional discrepancy~\cite{debortoli2021dsb,deng2024reflectedSB}. At terminal time (Fig.~\ref{fig:csb}), the SB objective yields a direct cross-modal consistency score that filters relevant evidence and, in turn, guides compute-budget allocation across iterations. Viewed as a diffusion-style decoder with a controllable step count, SB further enables step-adaptive fusion of the filtered cross-modal evidence. Our contributions are:

\begin{itemize}
\item \textbf{Consistency-as-Transport.} Estimate cross-modal consistency via a Schrödinger bridge that transports between endpoint distributions, yielding a calibrated reachability while eliminating auxiliary alignment networks.
\item \textbf{Coupling as Bottleneck.} A static, entropy-regularized optimal-transport coupling with evidence screening serves as a lightweight implicit interaction bottleneck, which unifying compute budget allocation, suppressing noise propagation, and mitigating negative transfer under modality imbalance and asynchrony.
\item \textbf{Cascaded SB Localizer.} Few-step bridges propose coarse intervals, followed by an evidence-guided, step-adaptive refiner with time mapping that targets cross-modal inconsistency for precise event localization. To our knowledge, IaMSB is the first method to apply the diffusion model to audio-visual deepfake localization.
\end{itemize}

Finally, IaMSB addresses issues \textbf{(i)–(iii)} by limiting cross-modal noise, adaptively reallocating compute toward suspicious modalities, and coupling an $\mathcal{O}(1)$ consistency score with an $\mathcal{O}(T)$ refinement bridge for finer temporal modeling. On audio–visual deepfake localization benchmarks, it achieves precise interval localization, robust cross-modal consistency, and improves AP@0.95 by $3\%\sim10\%$.

\section{Related Work}
\noindent\textbf{Audio–Visual Representation Learning.} A/V representations have evolved from masked unit prediction on speech–vision pairs to joint contrastive–masked objectives that explicitly align temporal granularity across streams \cite{shi2022avhubert,gong2023cavmae,huang2023mavil}. Early masked clustering supplies lip–speech priors \cite{shi2022avhubert}. Contrastive MAE formulations then strengthen cross-modal invariances and transferability \cite{gong2023cavmae,huang2023mavil}. Recent refinements decouple reconstruction from alignment and target fine-grained synchronization or tri-modal guidance, improving downstream consistency reasoning and interval sensitivity \cite{sun2024hicmae,araujo2025cavmaesync,ishikawa2025lgcavmae}.

\noindent\textbf{Audio–Visual Deepfake Localization.} BA-TFD \cite{cai2022lavdf_dicta} follows a content-driven two-stage paradigm that generates proposals from audio–visual cues and applies lightweight temporal refinement to yield interval predictions; BA-TFD+ \cite{cai2023lavdf_cviu} extends this design with stronger heads and denser supervision to sharpen boundaries and stabilize training. UMMAFormer \cite{zhang2023ummaformer} employs a multimodal-adaptive architecture that modulates fusion under reconstruction-style perturbations, improving boundary sensitivity while retaining robustness to cross-stream asynchrony. DiMoDif \cite{koutlis2024dimodif} explicitly models discourse-level inconsistencies beyond lip-sync, leveraging semantic conflicts to refine hypotheses and suppress spurious intervals. RegQAV \cite{zhu2025regqav} adopts a query-based decoder with register-enhanced tokens, regularizing decoding to tighten boundaries and maintain high-IoU precision under a fixed proposal budget. Other baselines include MMMS-BA \cite{katamneni2024mmmsba}, which uses strong cross-modal attention that boosts discrimination but can propagate noise under modality imbalance, and CFPRF \cite{wu2024cfprf}, an audio-side coarse-to-fine proposal refinement that stabilizes hypotheses and runtime. Complementary resources stress longer clips and harder partial forgeries \cite{avdeepfake1m2024,liu2024lipslying}.

\noindent\textbf{Diffusion and Schr\"odinger Bridges for Event Boundaries.} Generative decoders offer step-controllable inference and traceable denoising trajectories for boundary discovery \cite{xu2024denoiseLoc,wang2024fas,hwang2025diffgebd,chen2025diffdvc}. Beyond score-based diffusion, Schr\"odinger Bridges provide entropy-regularized transport between prior and posterior distributions, enabling calibrated progress measures that naturally couple detection confidence with compute budgeting \cite{debortoli2021dsb,deng2024reflectedSB}.

\begin{figure*}[t]
    \centering
    \includegraphics[width=1\linewidth]{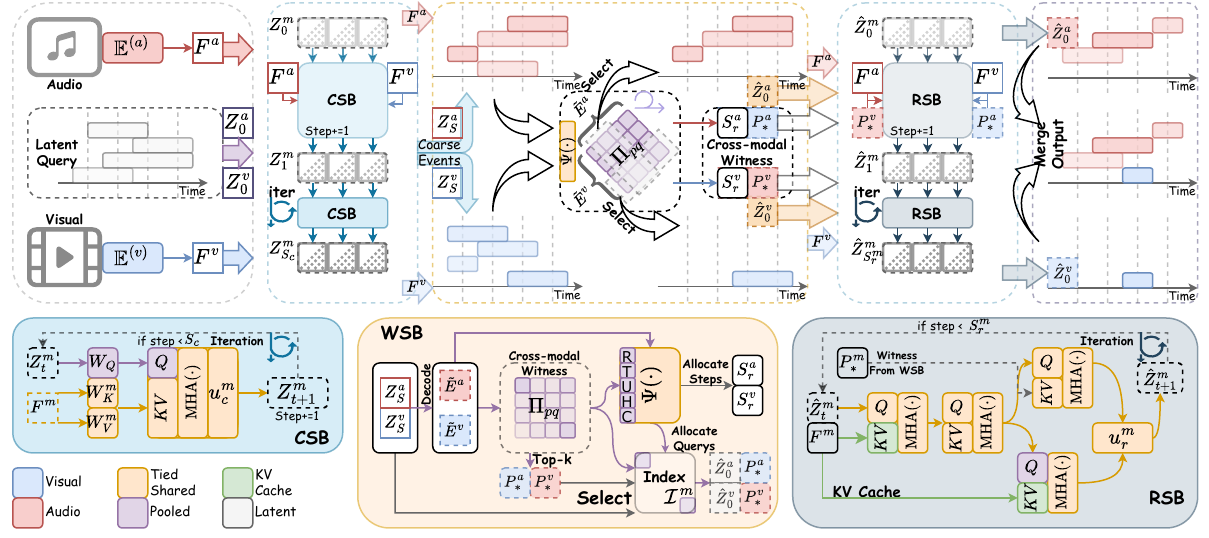}
    \caption{Overview of IaMSB. The extractors $\mathbb{E}^a$ and $\mathbb{E}^v$ produce modality–specific token sequences. The Coarse/ Witness/ Refinement Schrödinger Bridges (CSB/WSB/RSB) are described in Secs.~\ref{sec:csb}, \ref{sec:wsb}, and \ref{sec:rsb}. Heads $\mu_c^m,\mu_r^m$ follow Eq.~\eqref{eq:FFN}; $\mathrm{MHA}(\cdot)$ follows Eq.~\ref{eq:MHA}, with $Q$/$K$/$V$ roles indicated. The coarse stage yields $N_{\mathrm{ev}}$ latent events; after WSB screening they reduce to $\hat N^a$ and $\hat N^v$.}
    \vskip -1em
    \label{fig:placeholder}
\end{figure*}

\section{Method}
This section introduces IaMSB (Inconsistency-aware Multimodal Schrödinger Bridge), summarizes Schrödinger Bridge (SB) preliminaries that ground the formulation and fix notation, and then presents a cascaded bridge architecture. As shown in Fig.~\ref{fig:placeholder}, IaMSB consists of: (i) Coarse SB (CSB) proposes per-modality interval candidates via few-step, minimal-compute updates; (ii) Witness SB (WSB) selects witnesses and gates fusion through a lightweight latent bottleneck, allocates events and steps; (iii) Refinement SB (RSB) refines selected queries with step-tunable updates, injects witnesses, and operates under a unified budget.

\subsection{Problem Definition and Formulation}
For the $i$-th sample, the modality-specific backbones output feature tensors $F^{a}\in\mathbb{R}^{L_i^{a}\times C}$ and $F^{v}\in\mathbb{R}^{L_i^{v}\times C}$ for audio ($a$) and visual ($v$) modalities, respectively, with per-token time granularities $\Delta t^{a}$ and $\Delta t^{v}$. Final outputs are normalized to intervals on $[0,1]$ and mapped to absolute time. 

\noindent
\textbf{Problem Definition.} For localization, the model outputs, for the $k$-th event in modality $m$, the normalized start time $s^{m}_{i,k}\in[0,1]$, the event duration $\ell^{m}_{i,k}\in(0,1]$, and the confidence score $\pi^{m}_{i,k}\in(0,1)$, 
where $K^{m}_i$ denotes the predefined number of candidate events. Start and end times are
\begin{equation}
\begin{aligned}
t^{m}_{i,\mathrm{st}}&=\operatorname{clip}\!\big(s^{m}_{i,k}\,L^{m}_i\,\Delta t^{m},\,0,\,L^{m}_i\,\Delta t^{m}\big),\\
t^{m}_{i,\mathrm{ed}}&=\operatorname{clip}\!\big((s^{m}_{i,k}+\ell^{m}_{i,k})\,L^{m}_i\,\Delta t^{m},\,0,\,L^{m}_i\,\Delta t^{m}\big),
\end{aligned}
\label{eq:map_abs}
\end{equation}
where $\operatorname{clip}(x,a,b)=\min(\max(x,a),b)$, $L^{m}_i\,\Delta t^{m}$ is absolute time duration (second). 

\noindent
\textbf{Formulation.}
Let $m\in\{a,v\}$. For each stage, $F^m$ serves as the per-stage memory bank. Queries are initialized from latent codes $Z^m_0\in\mathbb{R}^{N_{\mathrm{ev}}\times C}$ with $N_{\mathrm{ev}}$ event tokens, follow 
\begin{itemize}
    \item \textbf{CSB} runs for $S_c$ steps, yielding $Z^m_t$ at the $t$-th step ($t=1,\dots,S_c$) and the coarse output $Z^m_{S_c}$ denote as $Z^m_\mathrm{out}$.
    \item \textbf{WSB} computes cross-modal witnesses $P^m$, and allocates the unified step budget $S_{\mathrm{tgt}}$ and event budget $N_{\mathrm{ev}}$ across modalities, resulting in per-modality refinement steps $S_r^a, S_r^v$ and selected counts $\hat N^a,\hat N^v$. WSB further produces an index matrix $\mathcal{I}^m$ that selects $\hat N^m$ events from $Z^m_\mathrm{out}$ and their corresponding witnesses from $P^m$, forming the masked subsets $\hat Z^m_\mathrm{out}$ and $\hat P^m$ for the refinement. 
    \item \textbf{RSB} runs for $S^m_r$ steps, yielding $\hat Z^m_t$ at the $t$-th step ($t=1,\dots,S^m_r$), the final latent sequence is $\hat Z^m_\mathrm{out}\in\mathbb{R}^{N^{m}\times C}$. 
    \item \textbf{Output.} For modality $m$, define $H^m(\cdot)$ as a modality-specific task head (FC layer) that maps $\hat Z^m_{out}$ to events set $E^m$. Each token is mapped to the start time $s$, duration $\ell$ and confidence score $\pi$, use $\phi(x)=\frac{1}{2}(1+\tanh x)$ limit the value to $[0, 1]$ and mitigate edge saturation.
\end{itemize}

\subsection{Preliminaries: Schrödinger Bridge}

Let $\Omega$ be a continuous path space and let the reference measure $R$ be induced by a Wiener process $W_t$ with noise intensity $\sigma>0$, $t$ is time step. Given \textbf{endpoint} marginals $\nu_0$ and $\nu_1$, the Schrödinger Bridge solves
\vskip -1.5em
\[
\min_{Q}\ \mathrm{KL}(Q\|R)\quad \text{s.t.}\quad Q_0=\nu_0,\ Q_1=\nu_1,
\]
\vskip -0.5em
\noindent
where $Q$ is trajectory. An equivalent controlled diffusion is
\vskip -1.5em
\[
\mathrm{d}Z_t=u_t(Z_t)\,\mathrm{d}t+\sigma\,\mathrm{d}W_t,\quad
\mathcal{E}(Q)=\mathbb{E}_Q\!\int_{0}^{1}\frac{\|u_t(Z_t)\|_2^2}{2\sigma^2}\,\mathrm{d}t ,
\]
\vskip -0.5em
\noindent
where $u_t(\cdot)$ is the control drift and $\mathcal{E}(Q)$ is the expected control energy. Partition $[0,1]$ into $S\in\mathbb{N}$ steps. Conditioning on $X$, define the discrete propagator
\[
Z_0\sim \nu_0,\qquad Z_S=\Phi_S(Z_0\,|\,X)\approx \nu_1(\cdot\,|\,X) .
\]
Given tolerance $\varepsilon>0$ and a discrepancy $D$, define the minimum step count
\[
S^\star(\varepsilon)=\min\Big\{S\in\mathbb{N}:\ D\big(\Phi_S(\nu_0\,|\,X),\ \nu_1(\cdot\,|\,X)\big)\le \varepsilon\Big\}.
\]
In this work $\nu_0$ encodes a sparse interval prior (e.g., localization result) and $\nu_1(\cdot\,|\,X)$ encodes an observation aligned posterior (e.g., this modality or another's ground truth). The bridge advances $\nu_0$ to $\nu_1$ through $\Phi_S$ while evolving latent states $Z_t$ into concrete interval predictions via the task head. We use the pair $\big(\mathcal{E}(Q),\,S^\star(\varepsilon)\big)$ as an explicit scale for cross-modal resource allocation and interaction.

\subsection{Localization and Budget Optimization}
This subsection presents network structure and three stage (CSR, WSB, RSB) localization under a unified budget.

\subsubsection{Model Architecture}
The bridge uses standard multi head attention and a SwiGLU feed forward network. Let $Q\in\mathbb R^{B\times L_q\times C}$ and $K,V\in\mathbb R^{B\times L_k\times C}$, with $H$ heads and per head dimension $d=C/H\in\mathbb Z^+$. With weights $W_q,W_k,W_v,W_o\in\mathbb R^{C\times C}$ split into $W_{q,h},W_{k,h},W_{v,h}\in\mathbb R^{C\times d}$ for $h=1,\dots,H$, we have $\mathrm{MHA}(Q,K,V) =$
\vskip -1.5em
\begin{equation}
\label{eq:MHA}
\Big[\bigoplus_{h=1}^{H}\mathrm{softmax}\Big(\frac{(QW_{q,h})(KW_{k,h})^{\top}}{\sqrt{d}}\Big)(VW_{v,h})\Big] W_o.
\end{equation}
\vskip -0.25em
\noindent
Unless stated otherwise, these weights are intrinsic to the structure and not shared across modules. With intermediate dimension $d_{\mathrm{ff}}=\alpha C$ (in this paper, $\alpha=8/3$) and weights $W_1,W_2\in\mathbb{R}^{C\times d_{\mathrm{ff}}}$ and $W_3\in\mathbb{R}^{d_{\mathrm{ff}}\times C}$, define
\vskip -0.75em
\begin{equation}
\mathrm{FFN}(x)=\big(\mathrm{SiLU}(xW_1)\odot(xW_2)\big)\,W_3,\,
\label{eq:FFN}
\end{equation}
\vskip -0.25em
\noindent
where $\mathrm{SiLU}(u)=u/(1+e^{-u})$. 

\subsubsection{Coarse Schrödinger Bridge (CSB)}
\label{sec:csb}
CSB adopts a lightweight design that strengthens temporal coherence and produces coarse event proposals to drive cross-modal interaction and budget allocation. Let $Z^{m}_{t}\in\mathbb{R}^{N_{\mathrm{ev}}\times C}$ be the latent sequence at step $t$, $F^{m}\in\mathbb{R}^{L^{m}\times C}$ is the conditional observation features, and $N_{\mathrm{ev}}$ the maximum candidate count. Coarse updates read features via cross attention and advance with a residual step for a fixed $S_{c}$ steps per modality, producing $N_{\mathrm{ev}}$ candidates via
\[
\begin{aligned}
U^{m}_{t}&=\mathrm{LN}\!\Big(Z^{m}_{t}+\mathrm{MHA}\big(Z^{m}_{t},\,F^{m},\,F^{m}\big)\Big),\\
Z^{m}_{t+1}&=Z^{m}_{t}+\Delta\,u^{m}_{\mathrm{c}}(U^{m}_{t}),\qquad u^{m}_{\mathrm{c}}(X)=\mathrm{FFN}(X),
\end{aligned}
\]
where $W_{q}\in\mathbb{R}^{C\times C}$ in $\mathrm{MHA}$ is a shared query map and the rest are modality specific, $\mathrm{LN}$ is layer normalization, $u^{m}_{\mathrm{c}}$ approximates a continuous time drift at coarse scale, $\mathrm{FFN}$ is implemented as in \eqref{eq:FFN}, and $\Delta=1/S_{c}>0$ is the step size. 
The control energy is
\vskip -0.25em
\begin{equation}
\mathcal E^{m}_{\mathrm{prior}}
=\sum_{t=0}^{S_c-1}\frac{\Delta\,\|u^{m}_{\mathrm{c}}(U^{m}_{t})\|_2^2}{2\sigma^2},\qquad \sigma>0.
\label{eq:prior_energy}
\end{equation}
\vskip -0.25em
\noindent
Denote output $Z^{m}_\mathrm{out} = Z^{m}_{S_{c}}$, using head $H^{m}$ (stop grad.) on $Z^{m}_\mathrm{out}$ yields coarse candidates $\tilde E^{m}=\{(s^{m}_{n},\,\ell^{m}_{n},\,\pi^{m}_{n})\}_{n=1}^{N_{\mathrm{ev}}}$. The complexity of CSB is $\mathcal{O}(N_{\mathrm{ev}}L^{m})$, where $L^{m}$ is the number of temporal tokens in modality $m$; it is more efficient than standard cross-attention~\cite{zhang2023ummaformer, yang2023avoiddf} with complexity $\mathcal{O}(L^{a}L^{v})$ and does not sacrifice the temporal resolution of any modality. CSB provides coarse events at a lower cost.

\subsubsection{Witness Schr\"odinger Bridge (WSB)}\label{sec:wsb}
WSB performs a static SB mapping on coarse events from the counterpart modality to drive budget allocation and establish a bottleneck for cross-modal latent interaction. For intervals \(I^{a}_{p}=[\,s^{a}_{p},\,s^{a}_{p}+\ell^{a}_{p}]\) and \(I^{v}_{q}=[\,s^{v}_{q},\,s^{v}_{q}+\ell^{v}_{q}]\), define $\mathrm{IoU}(I,J)=\frac{|I\cap J|}{|I\cup J|}$,
\vskip -1.25em
\[
\begin{aligned}
c_{pq}&=\lambda_{t}|s^{a}_{p}-s^{v}_{q}|
+\lambda_{\ell}|\ell^{a}_{p}-\ell^{v}_{q}|\\
&+\lambda_{o}\!\big(1-\mathrm{IoU}(I^{a}_{p},I^{v}_{q})\big)
+\lambda_{\pi}(\pi^{a}_{p}-\pi^{v}_{q})^{2},
\end{aligned}
\]
\vskip -0.25em
\noindent
$\lambda_t, \lambda_\ell, \lambda_o, \lambda_\pi$ are learnable parameters, for \(p,q\in\{1,\dots,N_{\mathrm{ev}}\}\). Let \(C=[c_{pq}]\in\mathbb{R}^{N_{\mathrm{ev}}\times N_{\mathrm{ev}}}\) and normalize confidences to marginals \(a,b\in\mathbb R^{N_{\mathrm{ev}}-1}\). The entropy-regularized OT (a static SB) coupling is
\vskip -0.25em
\[
\begin{aligned}
&\Pi=\arg\min_{\Pi\ge0}\ \langle \Pi,C\rangle+\sum_{p,q}\Pi_{pq}\log \Pi_{pq}\\
&\text{s.t.}\quad
\Pi\mathbf{1}_{N_\mathrm{ev}}=a,\ \Pi^\top\mathbf{1}_{N_\mathrm{ev}}=b,
\end{aligned}
\]
\vskip -0.25em
\noindent
with \(\Pi\in\mathbb{R}^{N_{\mathrm{ev}}\times N_{\mathrm{ev}}}\). Define statistics
\vskip -1.25em
\[
\begin{aligned}
\resizebox{1.0\linewidth}{!}{$
\begin{aligned}
&R=\sum_{p,q}\Pi_{pq}\,c_{pq},\quad
T=\sum_{p,q}\Pi_{pq},\quad
U=1-T,\quad\tilde{\Pi}_{pq}=\frac{\Pi_{pq}}{T+\varepsilon},\\
&H=-\sum_{p,q}\tilde{\Pi}_{pq}\log(\tilde{\Pi}_{pq}+\varepsilon),\ 
C=1-\sum_{p,q}\Pi_{pq}\,\mathrm{IoU}(I^{a}_{p},I^{v}_{q}),
\end{aligned}
$}
\end{aligned}
\]
\vskip -0.25em
\noindent
where \(\varepsilon>0\) stabilizes normalization. Here \(R\) is a weighted residual, \(U\) an unmatched rate, \(H\) a normalized coupling entropy, and \(C\) an interval inconsistency rate.
Row-wise keep only the top-\(k\) entries of \(\tilde{\Pi}\) (zero the rest), yielding a \(k\)-sparse row structure. For each modality \(m\), collect the selected opposite latents into a witness set \(P^m\in\mathbb{R}^{N_{\mathrm{ev}}\times k\times C}\) (the cross-modal bottleneck).
Let \(\Psi^{m}=\Psi^{m}(Z^m_\mathrm{out})\) be a one-layer Transformer. Form tokens
\[
G=[\Psi^{a},\ \Psi^{v},\ \Psi^{a}-\Psi^{v},\ \Psi^{a}\odot\Psi^{v},\ W_G \Phi_w]\in\mathbb{R}^{5\times d_s},
\]
where $\Phi_w=[R,T,U,H,C]^\top$, $W_G\in\mathbb R^{5\times d_s}$ is learnable. Self-interact with standard self-attention and FFN, then, through Softplus activation and a 1-layer MLP, obtain $\hat{S}=[\hat{S}_{a\mid v},\hat{S}_{v\mid a}]$.
Combine with prior control energies (\(\mathcal E^{m}_{\mathrm{prior}}\) in~\eqref{eq:prior_energy}) to get cross-modal weights:
\[
z=W_z\,[\,\hat{S}_{a\mid v},\ \hat{S}_{v\mid a},\ \mathcal E^{v}_{\mathrm{prior}}-\mathcal E^{a}_{\mathrm{prior}}\,]^\top+b_z,
\]
get $(w^{v},w^{a})=\mathrm{softmax}(z)$. Subsequently, integer approximate segmentation is performed on the total event $N_\mathrm{ev}$ and the total step budget $S_\mathrm{tgt}$, allocate budget $S^m_r = 2\lfloor \tfrac{w^mS_\mathrm{tgt}}{2} + 0.5 \rfloor$, $\hat N^m = 2\lfloor \tfrac{w^mN_\mathrm{ev}}{2} + 0.5 \rfloor$, selected $\hat N^m$ token runs exactly \(S^{m}_r\) refinement iterations.

Next, we select cross-modal evidence, with \(r_p=\sum_q \Pi_{pq}\), \(c_q=\sum_p \Pi_{pq}\), define priorities
\vskip -1.25em
\[
\gamma^{(a)}_p=1-r_p,\quad \gamma^{(v)}_q=1-c_q,\quad
\pi^{m}_j=\frac{\exp(\gamma^{m}_j/\tau)}{\sum_{j'}\exp(\gamma^{m}_{j'}/\tau)}.
\]
\vskip -0.25em
\noindent
Sample \(\mathcal I^{m}\subset\{1,\dots,N_{\mathrm{ev}}\}\),  \(|\mathcal I^{m}|=\hat{N}^{m}\) without replacement by \(\pi^{m}\). The outputs to the refinement stage are $\tilde{Z}^{m}_\mathrm{out}=Z^{m}_\mathrm{out}[\mathcal I^{m},:]\in\mathbb{R}^{\hat{N}^{m}\times C}$, $\hat{P}^{m}=P^{m}[\mathcal I^{m},:,:]\in\mathbb{R}^{\hat{N}^{m}\times k\times C}$. WSB always interacts with events, thus being time-independent with a constant complexity of $\mathcal O(1)$.

\subsubsection{Refinement Schrödinger Bridge (RSB)}
\label{sec:rsb}
Next, we refine coarse-grained events and introduce cross-modal witnesses. Given the selected coarse-stage latent sequence $\tilde Z^{m}_\mathrm{out}\in\mathbb{R}^{\hat N^m\times C}$ (CSB), the witness tensor $\hat{P}^{m}\in\mathbb{R}^{\hat N^m\times k\times C}$, and the observation features $F^{m}\in\mathbb{R}^{L^{m}\times C}$, the refinement stage (RSB) evolves the iterates $\{\hat Z^{m}_t\}_{t=0}^{S_r^m}$ with initialization $\hat Z^{m}_0=\tilde Z^{m}_\mathrm{out}$; let $t\in\{0,\dots,S_r^m\}$ denote the refinement iteration index. The refinement iterations are
\begin{equation}
\label{eq:rsb1}
U^{m}_{t,1}=\mathrm{LN}\!\Big(\hat Z^{m}_{t}+\mathrm{MHA}_\mathrm{shr}\big(Z^{m}_{t},\,F^{m},\,F^{m}\big)\Big),
\end{equation}
\[
U^{m}_{t,2}=\mathrm{LN}\!\Big(U^{m}_{t,1}+\mathrm{MHA}\big(U^{m}_{t,1},\,U^{m}_{t,1},\,U^{m}_{t,1}\big)\Big),
\]
Here, $\mathrm{MHA}_\mathrm{shr}$ stands for shared $KV$. Inject opposing high confidence candidates as witnesses to shrink cross modal residuals and reach the target distribution in fewer steps. For each query $n\in\{1,\dots,\hat N^m\}$,
\[
R^{m}_{\mathrm{wit},n}=\mathrm{MHA}\!\big(U^{m}_{t,2}[n,:],\,\hat P^{m}[n,:,:],\,\hat P^{m}[n,:,:]\big)\in\mathbb{R}^{C},
\]
which stacks to $R^{m}_{\mathrm{wit}}\in\mathbb{R}^{N^m_*\times C}$. Merge memory to obtain 
\[
U^{m}_{t,3}=\mathrm{LN}\!\Big(U^{m}_{t,2}+R^{m}_{\mathrm{wit}}
+\mathrm{MHA}_{\mathrm{shr}}\big(U^{m}_{t,2},\,F^{m},\,F^{m}\big)\Big).
\]
where $\mathrm{MHA}_{\mathrm{shr}}$ is modality-shared $Q$ and share $KV$ with Eq.~\eqref{eq:rsb1}.This serves as the drift input. Advance with residual updates using step size $\Delta=1/S^m_r$ and drift
\[
Z^{m}_{t+1}=Z^{m}_{t}+\Delta\,u^{m}_{\mathrm{ref}}(U^{m}_{t,3}),\qquad
u^{m}_{\mathrm{ref}}(X)=\mathrm{FFN}(X),
\]
where $\mathrm{FFN}$ is as in \eqref{eq:FFN}, output $\hat Z^m_{S^m_r}$ denote as $\hat Z^m_\mathrm{out}$. Per time step \(t\), the generation of \(U^{m}_{t,1}\) and \(U^{m}_{t,3}\) incurs \(\mathcal{O}(\hat N^{m} L^{m})\) cost, \(U^{m}_{t,2}\) incurs \(\mathcal{O}\!\big((\hat N^{m})^{2}\big)\), and computing \(R^{m}_{\mathrm{wit},n}\) incurs \(\mathcal{O}(n_{k}\,\hat N^{m})\). All of these costs scale linearly with the temporal length.

\subsubsection{Decoding and Optimizing Objective}
Decode the RSB's output $\hat Z^m_\mathrm{out}$, updated candidates with
\[
\{(s^{m}_k,\ell^{m}_k,\pi^{m}_k)\}_{k=1}^{\hat N^m}=H^{m}(\hat Z^m_\mathrm{out}).
\]
Map $(s^{m}_k,\ell^{m}_k)$ to the absolute timeline via Eq.\eqref{eq:map_abs}. Fuse across modalities to obtain the final event predictions. The localization loss has matching, negative, and coverage terms with $\mathcal{M}^{m}$ the set of positive matches and $\mathcal{U}^{m}$ the set of unmatched candidates, $\mathcal L^{m}_{\mathrm{loc}}=$
\begin{equation}
\nonumber
\resizebox{0.9\linewidth}{!}{$
\begin{aligned}
&\sum_{(p,g)\in\mathcal M^{m}}\!\Big[
\bigl(1-\mathrm{EIoU}(p,g)\bigr)
+\mathcal H(p,g)
+\mathrm{BCE}(\pi^{m}_p,1)\Big]\\
&\quad+\sum_{k\in\mathcal U^{m}}\mathrm{BCE}(\pi^{m}_k,0)
+\sum_{g}\exp\!\bigl(-\beta \max_{p}\mathrm{IoU}(p,g)\bigr),
\end{aligned}
$}
\end{equation}
where $\mathcal H$ is huber operator, $\beta$ is hyper parameter. Margin-based ranking imposes identifiable, calibrated ordering on directional uncertainty, tracks true difficulty and steers budget to the harder side:
\vskip  -0.5em
\[
\mathcal L_{\mathrm{rank}}=\max\big(0,\ m_0-(\hat S_{a\mid v}-\hat S_{v\mid a})\big).
\]
\vskip  -0.25em
\noindent
where $m_0>0$ is the margin hyperparameter specifying the minimum separation between the directional scores.

With step granularity $\rho$ and selectable set $\mathcal S$, we require that adding an extra $\rho$-step should not worsen the objective. Let the eligible index set be $\mathcal J^{(m)}=\{\,j\in\mathcal I^{(m)}\mid s_j+\rho\in\mathcal S\,\}$ and define the average marginal degradation
\vskip -0.5em
\begin{equation}
\nonumber
\resizebox{\linewidth}{!}{$
\bar\Delta^{(m)}
=\frac{1}{\max\{1,|\mathcal J^{(m)}|\}}
\sum_{j\in\mathcal J^{(m)}}
\big[\ \kappa^{(m)}_j(s_j+\rho)-\kappa^{(m)}_j(s_j)\ \big]_+,
$}
\end{equation}
\vskip -0.25em
\noindent
where $\kappa^{(m)}_j(s)$ denotes the objective value of candidate $j$ at $s$ steps. The regularizer compares the per-step marginal effects across modalities with budget weights:
\[
\mathcal L_{\mathrm{svn}}
=\ \frac{\hat N^{a}S^{a}_{r}}{\rho}\,\big[\ \bar\Delta^{(v)}-\bar\Delta^{(a)}\ \big]_+
\ +\ \frac{\hat N^{v}S^{v}_{r}}{\rho}\,\big[\ \bar\Delta^{(a)}-\bar\Delta^{(v)}\ \big]_+.
\]
This form penalizes (i) violations of non-increase via the hinge on $\bar\Delta^{(m)}$, and (ii) cross-modal imbalance by weighting with the actual per-modality budget $\hat N^{m}S^{m}_{r}$.
Total loss
\vskip -1em
\[
\mathcal L=\sum_{m\in\{a,v\}}\mathcal L^{m}_{\mathrm{loc}}
+\lambda_{\mathrm{rank}}\mathcal L_{\mathrm{rank}}
+\lambda_{\mathrm{svn}}\mathcal L_{\mathrm{svn}},
\]
\vskip -0.5em
\noindent
where $\lambda_{\mathrm{rank}},\lambda_{\mathrm{svn}}>0$.

\begin{table*}[t]
\centering
\caption{Temporal localization on \textbf{LAV-DF} and \textbf{TVIL}. The best in each column is in \textbf{bold}.}
\resizebox{0.8\linewidth}{!}{%
\setlength{\tabcolsep}{8pt}
\begin{tabular}{l | l ccc cccc}
\toprule
\textbf{Dataset} & \textbf{Methods} & \textbf{AP@0.5} & \textbf{AP@0.75} & \textbf{AP@0.95} & \textbf{AR@10} & \textbf{AR@20} & \textbf{AR@50} & \textbf{AR@100} \\
\midrule
\multirow{11}{*}{LAV\text{-}DF}
 & MDS~\cite{chugh2020mds} & 12.78 & 1.62 & 0.00 & 32.15 & 36.71 & 34.39 & 37.88 \\
 & AVTFD~\cite{liu2024avtfd} & 94.89 & 74.87 & 1.94 & 71.24 & 72.17 & 74.24 & 76.12 \\
 & BA\text{-}TFD~\cite{cai2022lavdf_dicta} & 93.10 & 71.22 & 1.32 & 69.80 & 71.01 & 73.55 & 75.40 \\
 & BA\text{-}TFD+~\cite{cai2023lavdf_cviu} & 96.30 & 84.96 & 4.44 & 78.75 & 79.40 & 80.48 & 81.62 \\
 & ActionFormer~\cite{zhang2022actionformer} & 85.23 & 59.05 & 0.93 & 76.93 & 77.19 & 77.23 & 77.23 \\
 & UMMAFormer~\cite{zhang2023ummaformer} & 98.83 & 95.54 & 37.61 & 92.10 & 92.42 & 92.47 & 92.48 \\
 & MMMS\text{-}BA~\cite{katamneni2024mmmsba} & 97.56 & 95.25 & 39.02 & 89.42 & \textbf{95.93} & 93.45 & 94.05 \\
 & DiMoDif~\cite{koutlis2024dimodif} & 95.50 & 87.90 & 20.60 & 91.40 & 92.70 & 93.70 & 94.20 \\
 & RegQAV~\cite{zhu2025regqav} & 94.10 & 88.10 & 27.60 & 91.70 & 91.80 & 93.20 & 94.60 \\
 & \textbf{IaMSB (ours)} & \textbf{99.33} & \textbf{95.62} & \textbf{55.92} & \textbf{94.68} & 95.41 & \textbf{95.50} & \textbf{95.52} \\
\midrule
\multirow{6}{*}{TVIL}
 & BA\text{-}TFD+~\cite{cai2023lavdf_cviu} & 76.90 & 38.50 & 0.25 & 66.90 & 64.08 & 60.77 & 58.42 \\
 & ActionFormer~\cite{zhang2022actionformer} & 86.27 & 83.03 & 28.17 & 84.82 & 85.77 & 88.10 & 88.49 \\
 & UMMAFormer~\cite{zhang2023ummaformer} & 88.68 & 84.70 & 62.43 & 87.09 & 88.21 & 90.43 & 91.16 \\
 & MMMS\text{-}BA~\cite{katamneni2024mmmsba} & 96.87 & 81.33 & 28.43 & 88.61 & 87.83 & 90.47 & 92.91 \\
 & \textbf{IaMSB (ours)} & \textbf{96.89} & \textbf{85.33} & \textbf{65.62} & \textbf{90.05} & \textbf{90.82} & \textbf{92.00} & \textbf{93.10} \\
\bottomrule
\end{tabular}%
}
\vskip -0.5em
\label{tab:first_2}
\end{table*}

\section{Experiments}
This section evaluates the performance of IaMSB. Through comparative analysis with baseline methods, supported by computational complexity assessment and visualization results, it provides a comprehensive characterization of IaMSB’s behavior and performance. Supplementary materials provide more details.

\subsection{Setup}

\noindent
\textbf{Datasets.} We evaluate on \textbf{LAV-DF} \cite{cai2022lavdf_dicta,cai2023lavdf_cviu}, \textbf{AV-Deepfake1M} \cite{avdeepfake1m2024}, and \textbf{TVIL} \cite{zhang2023ummaformer}. LAV-DF is a widely used benchmark; AV-Deepfake1M stresses long-form and partial forgeries; TVIL is a unilateral forgery of the visual modality. For TVIL we use a visual resolution of $216{\times}120$ and an audio embedding rate of $540$\,Hz. For AV-Deepfake1M and LAV-DF we use $128^2$ resolution and $640$\,Hz audio embeddings.

\noindent
\textbf{Hardware and Implement Details.}
All localization experiments are run on 4$\times$V100 (32GB) GPUs (AV-DeepFake1M) + 16$\times$R9 9950X CPUs and 2$\times$RTX-4090 (24GB) GPUs + 12$\times$R9 5900X CPUs (LAV-DF, TVIL) with bf16 precision and AdamW~\cite{adamw} optimizer sharding. The batch size of each GPU is 8, which accumulates to an equivalent 128. $\lambda_\mathrm{rank}, \lambda_\mathrm{svn} = 0.2$. We use a learning rate of $2.5\times10^{-4}$ with a cosine warmup schedule, and train all models for 40 epochs. 
The coarse stage (CSB) uses 2 steps per modality with $N_\mathrm{ev}=n$, which is tied to the evaluation metric \textbf{AR@}$n$; the witness stage (WSB) applies a single Sinkhorn pass to compute cross-modal statistics and select top-$16$ (or top-$N_\mathrm{ev}$ if $N_\mathrm{ev}<16$) witnesses; the refinement bridge (RSB)'s step budget is fixed to $\mathcal S_\mathrm{tgt}{=}12$. The selectable set $\mathcal S = \{0, 2\}$ and step granularity $\rho = 2$.

Encoders are ViT-S for both branches, initialized from VideoMAE~\cite{videomae} (visual) and WavLM~\cite{WavLM} (audio) knowledge; token embeddings and positional encodings are retrained, and for VideoMAE we insert a frame-wise group of \texttt{[CLS]} tokens, while all other encoder weights are frozen. 

\begin{table*}[t]
\centering
\caption{Temporal localization on \textbf{AV-Deepfake1M}. The best in each column is in \textbf{bold}. Comparison results are from~\citet{zhu2025regqav}.}
\resizebox{0.9\linewidth}{!}{
\begin{tabular}{l | l cccc ccccc}
\toprule
\textbf{Dataset} & \textbf{Methods} & \textbf{AP@0.5} & \textbf{AP@0.75} & \textbf{AP@0.90} & \textbf{AP@0.95} & \textbf{AR@50} & \textbf{AR@30} & \textbf{AR@20} & \textbf{AR@10} & \textbf{AR@5} \\
\midrule
\multirow{7}{*}{AVDeepfake1M}
& BA\text{-}TFD~\cite{cai2022lavdf_dicta}               & 37.37 &  6.34 &  0.19 &  0.02 & 45.55 & 40.37 & 35.95 & 30.66 & 26.82 \\
& BA\text{-}TFD+~\cite{cai2023lavdf_cviu}              & 44.42 & 13.64 &  0.48 &  0.03 & 48.86 & 44.51 & 40.37 & 34.67 & 29.88 \\
& ActionFormer~\cite{zhang2022actionformer}            & 36.08 & 12.01 &  1.23 &  0.16 & 27.11 & 27.08 & 27.00 & 26.60 & 25.80 \\
& UMMAFormer~\cite{zhang2023ummaformer}                & 51.64 & 28.07 &  7.65 &  1.58 & 44.07 & 43.93 & 43.45 & 42.09 & 40.27 \\
& DiMoDif~\cite{koutlis2024dimodif}                     & 86.93 & 75.95 & 28.72 &  5.43 & 81.57 & 80.85 & 80.25 & 78.84 & 76.64 \\
& RegQAV~\cite{zhu2025regqav}                           & 90.24 & 81.86 & 41.98 & 12.57 & 88.14 & 87.68 & 87.27 & 86.95 & 85.97 \\
& \textbf{IaMSB (ours)}                                 & \textbf{90.31} & \textbf{82.03} & \textbf{45.15} & \textbf{23.01} & \textbf{90.02} & \textbf{88.61} & \textbf{88.05} & \textbf{87.74} & \textbf{86.03} \\
\bottomrule
\end{tabular}}
\vskip -1em
\label{tab:avdf1m_main}
\end{table*}

\subsection{Comparison on LAV-DF and TVIL}
\textbf{Metric.}
We compared with the baseline on LAV-DF and TVIL datasets, and the results are shown in Tab.~\ref{tab:first_2}.
We report AP@$\{0.5,0.75,0.95\}$ and AR@$\{10,20,50,100\}$~\cite{cai2023lavdf_cviu, zhang2023ummaformer} on LAV-DF/TVIL. The experiment results of the comparison method are an official report or from~\citet{cai2023lavdf_cviu}.

\textbf{On LAV-DF}, Proposal–refine baselines such as BA-TFD/BA-TFD$+$~\cite{cai2022lavdf_dicta,cai2023lavdf_cviu} and anchor-free ActionFormer~\cite{zhang2022actionformer} provide strong recall; fusion-oriented UMMAFormer~\cite{zhang2023ummaformer} and MMMS-BA~\cite{katamneni2024mmmsba} strengthen cross-modal coupling; DiMoDif~\cite{koutlis2024dimodif} models discourse-level inconsistency; RegQAV~\cite{zhu2025regqav} stabilizes query decoding. In the table, several methods reach similar AR yet differ markedly at AP@0.95, indicating boundary accuracy is sensitive to where the limited refinement is placed.

IaMSB attains markedly higher performance at strict overlap (AP@0.95) than the baselines, suggesting that its temporally linear-complexity fusion yields finer boundary resolution and directly supports our central claim. At AR@20, the uniform choice of top-$k=16$ likely introduces redundancy within the cross-modal bottleneck, which can dampen recall. This observation further underscores the necessity of selective evidence routing rather than broad, indiscriminate exchange.

\textbf{TVIL} is a visual-only forgery setting (audio clean, manipulations in V as introduced with the benchmark in~\cite{zhang2023ummaformer}); Methods that emphasize visual refinement or query regularization, including ActionFormer~\cite{zhang2022actionformer}, UMMAFormer~\cite{zhang2023ummaformer}, RegQAV~\cite{zhu2025regqav}, and MMMS-BA~\cite{katamneni2024mmmsba}, the last of which incorporates a dedicated lip-movement sub-network for visual modality, achieve strong relaxed-IoU precision (AP@0.5), where visual boundary modeling dominates.

However, in a stricter IoU scenario, this gap is significantly magnified. The cross-attention of the three views enables MMMS-BA to make significant trade-in timing density. In contrast, UMMAFormer, which has a relatively lower cost of timing calculation, performs stably, while IaMSB still maintains the best performance. Because IaMSB allows for adaptive iteration, it consistently leads in the recall metric within a limited number of event tokens.

On LAV-DF, pronounced cross-modal inconsistencies make transport-derived cues informative; thus non-uniform step allocation yields clear AP@0.95 gains at comparable AR. On TVIL (visual-only tampering), cross-modal evidence is weak, so budget placement chiefly boosts recall and mid-range IoU, while strict-IoU tightening benefits more from visual-specialized decoders. For long clips, attention over sequences of length $S\approx LT$ scales as $\mathcal{O}(S^{2})$ and dominates compute; consequently, aggressive temporal downsampling or surrogate views can aid coarse discrimination yet often erode boundary precision at high IoU. IaMSB mitigates this trade-off by concentrating refinement where transport residuals are large, rather than distributing steps uniformly across time or modalities.

\subsection{Comparison on AV-Deepfake1M}
\noindent\textbf{Metrics.} AV-Deepfake1M follows an official temporal localization protocol that reports AP at multiple IoU thresholds (0.50/0.75/0.90/0.95) together with AR under different proposal budgets (50/30/20/10/5). 

In Table~\ref{tab:avdf1m_main}, proposal–refine and fusion-heavy methods show moderate AR but lower AP at stricter IoUs: BA-TFD~\cite{cai2022lavdf_dicta} and BA-TFD$+$~\cite{cai2023lavdf_cviu} rely on content-driven proposals with uniform refinement; ActionFormer~\cite{zhang2022actionformer} offers strong anchor-free proposals yet lacks target-aware boundary tightening in this setting; UMMAFormer~\cite{zhang2023ummaformer} strengthens cross-modal coupling but still spreads compute evenly across segments. Recent designs tuned for fine localization perform better across columns: DiMoDif~\cite{koutlis2024dimodif} improves both strict-IoU AP and AR, and RegQAV~\cite{zhu2025regqav} further raises all metrics under the same protocol by stabilizing query decoding with register-enhanced tokens.

Under the same evaluation, IaMSB improves strict-IoU AP while keeping AR competitive or higher across proposal budgets, yielding the best columns overall in Table~\ref{tab:avdf1m_main}. 
We observe the same pattern: coarse-grained localizers (e.g., temporal striding~\cite{zhang2023ummaformer}, aggressive temporal downsampling~\cite{zhu2025regqav}, or surrogate text-level alignment~\cite{koutlis2024dimodif}) are severely limited under strict-IoU (AP@0.90 or 0.95). This reinforces the need for a low-cost, budget-allocatable fusion architecture whose complexity is linear in time (or effectively time-agnostic), enabling fine-grained temporal decisions without prohibitive compute. Combine with the results in Tab.~\ref{tab:first_2}, it indicates the necessity of time resolution for high strict-IoU, as mentiond in Sec.~\ref{sec:intro}, issue \textbf{(iii)}.

\begin{figure*}[t]
  \centering
  \captionsetup[subfigure]{justification=centering}
  \begin{subfigure}[b]{0.33\linewidth}
    \includegraphics[width=\linewidth]{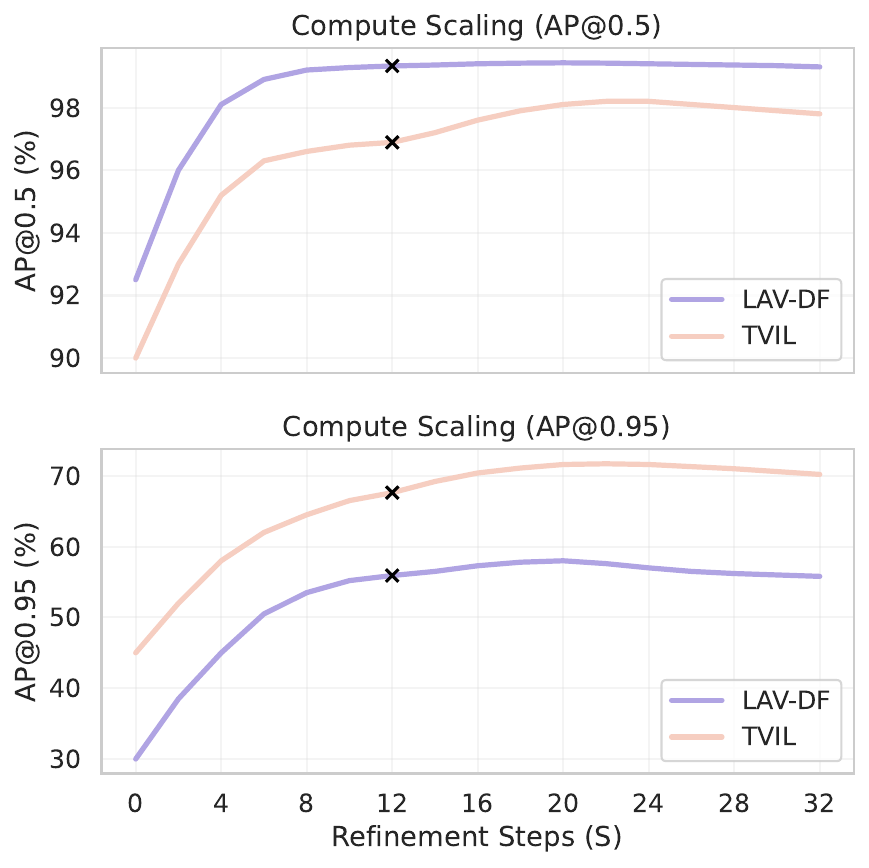}
    \subcaption{Compute scaling under a unified budget.}
    \label{fig:compute_scaling}
  \end{subfigure}
  \hfill
  \begin{subfigure}[b]{0.66\linewidth}
    \includegraphics[width=\linewidth]{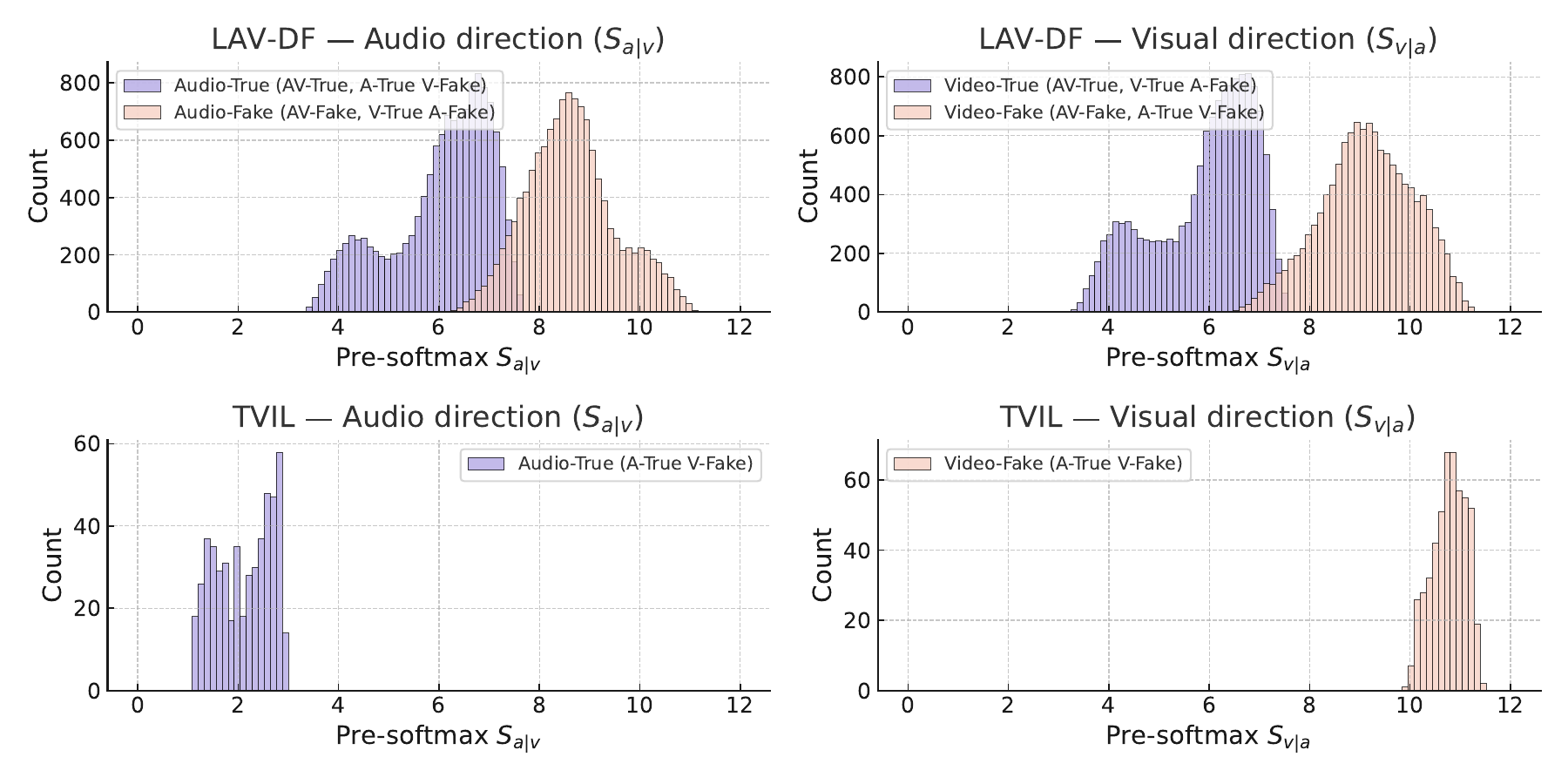}
    \subcaption{Directional scale tensors before softmax, map x-axis to RSB step.}
    \label{fig:sm_mosaic}
  \end{subfigure}
  \caption{Compute scaling and directional scales. (a) Step–accuracy trade-off of IaMSB under a fixed budget (sensitive of $S_\mathrm{tgt}$). (b) Pre-softmax directional scales that drive cross-modal allocation, with LAV-DF exhibiting overlapped True/False groups and TVIL showing the more extreme A-True V-Fake pattern. }
  \label{fig:scaling_and_scales}
    \vskip -1em
\end{figure*}

\subsection{Ablation and Resource Analyze}

\textbf{Structural.} As shown in Tab.~\ref{tab:struct-rev}, we evaluate four variants under the official LAV\text{-}DF protocol. Removing the coarse bridge (-CSB) mainly hurts recall due to missing proposals under the fixed extractor; using only CSB amplifies modality-imbalance errors and reduces high\text{-}IoU precision; dropping refinement (-RSB) preserves mid-range AP but loses boundary polish. Removing WSB eliminates cross-modal filtering and budget allocation, leading to a substantial degradation across AP/AR; this indicates that the cross-modal bottleneck and budgeting are necessary, and that each component’s functional role behaves as intended.

\begin{table}[t]
\caption{Structural ablations on LAV-DF. \checkmark means bridges kept. * means CSB forward 2 times and per modality RSB forward 6 times. \textbf{Bold} is our default setting.}
\label{tab:struct-rev}
\centering
\resizebox{\columnwidth}{!}{
\begin{tabular}{lll|ccc|cccc}
\toprule
\multicolumn{3}{c|}{Bridges} & \multicolumn{3}{c|}{AP} & \multicolumn{4}{c}{AR} \\
CSB & WSB & RSB & @0.5 & @0.75 & @0.95 & @10 & @20 & @50 & @100 \\
\midrule
\checkmark & \checkmark & \checkmark & \textbf{99.33} & \textbf{95.62} & \textbf{55.92} & \textbf{94.68} & \textbf{95.41} & \textbf{95.50} & \textbf{95.52} \\
           & \checkmark & \checkmark & 83.25 & 80.37 & 32.51 & 87.83 & 88.62 & 89.05 & 89.57 \\
\checkmark &            &            & 86.33 & 78.56 & 22.07 & 85.25 & 86.01 & 87.16 & 87.43 \\
\checkmark$^*$ &            & \checkmark$^*$ & 96.35 & 87.08 & 28.16 & 89.83 & 90.78 & 91.39 & 92.04 \\
\checkmark & \checkmark &            & 86.75 & 83.23 & 23.39 & 85.77 & 87.05 & 87.52 & 87.97 \\
\bottomrule
\end{tabular}}
\end{table}

\noindent
\textbf{Coarse-bridge sensitivity ($S_c$).} As shown in Tab.~\ref{tab:sc-rev}, under a fixed refinement budget for RSB, increasing the number of CSB steps yields the most pronounced gain when moving from \(S_c=1\) to \(S_c=2\), likely because the coarse stage provides tighter initial intervals; beyond \(S_c=2\), the observed differences are within experimental variance. Since CSB is designed to furnish WSB with a coarse yet sufficiently informative candidate set, \(S_c=2\) meets this objective and is adopted as our default.

\begin{table}[t]
\caption{Sensitivity to coarse steps $S_c$ under a fixed budget. \textbf{Bold} is our default setting.}
\label{tab:sc-rev}
\centering
\resizebox{\columnwidth}{!}{
\begin{tabular}{c|ccc|cccc}
\toprule
$S_c$ & AP@0.5 & AP@0.75 & AP@0.95 & AR@10 & AR@20 & AR@50 & AR@100 \\
\midrule
1 & 98.51 & 93.32 & 52.87 & 93.76 & 94.23 & 94.45 & 94.58 \\
2 & \textbf{99.33} & \textbf{95.62} & \textbf{55.92} & \textbf{94.68} & \textbf{95.41} & \textbf{95.50} & \textbf{95.52} \\
3 & 99.35 & 95.54 & 55.98 & 94.82 & 95.55 & 95.63 & 95.68 \\
\bottomrule
\end{tabular}}
\end{table}

\noindent
\textbf{Witness Top-$k$ selection.}
Tab.~\ref{tab:topk-rev} reports performance under different top-$k$ settings ($k\!\in\!\{2,\ldots,64\}$). The results show that more interaction is not necessarily better: a narrow bottleneck (small $k$) can under-expose cross-modal evidence, whereas an overly wide bottleneck (large $k$) weakens selectivity and admits noise. These trends support our weak-interaction bottleneck, which reduces WSB compute while improving accuracy. Together with the AR@20 difference on LAV-DF (Tab.~\ref{tab:first_2}), this result suggests that fusion can either be improved or noise can be introduced, as mentiond in Sec.~\ref{sec:intro}, issue \textbf{(i)},  selective interaction is a better solution.

\begin{table}[t]
\caption{Sensitivity to witness memory Top-$k$.}
\label{tab:topk-rev}
\centering
\resizebox{0.8\columnwidth}{!}{
\begin{tabular}{c|cccccc}
\toprule
 & \multicolumn{6}{c}{Top-$k$} \\
\cmidrule(lr){2-7}
$\mathrm{AP}@0.95$ & 2 & 4 & 8 & 16 & 32 & 64 \\
\midrule
 & 50.98 & 52.31 & 54.93 & \textbf{55.92} & 54.35 & 53.97 \\
\bottomrule
\end{tabular}
}
\vskip -0.5em
\end{table}

\noindent
\textbf{Cost.} We report two representative inference settings (Tab.~\ref{tab:resource}): under \emph{avg} (video $\mathbb{R}^{1\times 225\times 384}$, audio $\mathbb{R}^{1\times 225\times 384}$, equal allocation), per-step costs are \textbf{0.428} GFLOPs (CSB per step), $3\times 10^{-5}$ GFLOPs (single OT iteration), and \textbf{1.01} GFLOPs (RSB per step); under \emph{max} (video $\mathbb{R}^{1\times 750\times 384}$, audio $\mathbb{R}^{1\times 750\times 384}$, all refinement on one side), they are \textbf{0.739} GFLOPs, $3\times 10^{-5}$ GFLOPs, and \textbf{2.22} GFLOPs. Totally $\sim27.1$ GFLOPs, with 20ms for IaMSB, 100ms for extractor per-sample on single RTX4090. We admit that the computational cost brought by the backbone network using $\mathcal O(T^2)$ (ViTs) cannot be ignored. However, the analysis of the computational load of each part also indicates the linear correlation of the sample duration $T$ of the cascade bridge.

\begin{table}[t]
\centering
\caption{Resource requirements. "max." means $30s$ video; "avg." meas $\sim8.6s$ video, which is the average length of LAV-DF.}
\label{tab:resource}
\resizebox{\linewidth}{!}{
\begin{tabular}{lccc}
\hline
Method & Total Params [M] & Trainable Params [M] & FLOPs [G] \\
\hline
BA-TFD            & 5.5   & 5.5   & 948.1  \\
BA-TFD+           & 152.9 & 152.9 & 218.2  \\
UMMAFormer        & 165.9 & 49.7  & 1,562.9\\
RegQAV            & 116.5 & 19    & 251.1  \\
IaMSB(ours, avg.) & 70.6  & 10.3   & 93.0   \\
IaMSB(ours, max.) & 70.6  & 10.3   & 311.2  \\
\hline
\end{tabular}
}
\vskip -1em
\end{table}

\subsection{Visualization}

\noindent
\textbf{Cost scaling (Fig.~\ref{fig:compute_scaling}).}
Under a unified budget, increasing the number of refinement steps $S$ yields a monotonic improvement of AP across operating points, with clear diminishing returns at higher $S$. The curves show consistent gains on both LAV\text{-}DF and TVIL, and the separation between AP@\,$0.5$ and AP@\,$0.95$ highlights how extra steps increasingly benefit boundary “polishing” rather than coarse detection—evidence that compute is preferentially spent on hard, high-IoU cases. This visualization operationalizes a practical knob: choose $S$ to target either higher precision at strict IoU or faster throughput with minimal loss, but excessively high values yield diminishing returns.

\noindent
\textbf{Directional scales (Fig.~\ref{fig:sm_mosaic}).}
The histograms of pre-softmax scales $(S_{a\mid v}, S_{v\mid a})$ reveal dataset-dependent asymmetries in cross-modal evidence. On LAV\text{-}DF, True/False groups substantially overlap, indicating frequent near-symmetric interactions between modalities (e.g., benign asynchrony and short, subtle forgeries), whereas on TVIL the video-directed scale $S_{v\mid a}$ is more skewed in the A-True/V-Fake regime, reflecting stronger reliance on visual refinement when audio offers reliable “witness” cues. These patterns motivate our step allocator to assign more budget toward the direction with larger pre-softmax scale, while preserving robustness when distributions overlap.

\textbf{Strikingly}, reallocation arises even when the audio stream is genuine and supervision offers no incentive to weight audio: transport statistics still assign a non-zero quota to audio. This indicates that compute placement should be adaptive rather than modality-fixed. Uniform allocation is not necessarily optimal, and a clean modality can still aid temporal alignment and suppress false positives without increasing the total budget. Consequently, IaMSB can roughly discriminate authenticity in latent space and direct steps toward suspicious modalities under a unified budget. The evidence highlights the prevalence of compute-placement sensitivity and motivates step-wise asynchronous fusion, as mentiond in Sec.~\ref{sec:intro}, issue \textbf{(ii)}.

\section{Conclusion}
We present IaMSB, a budgeted two-stage Schrödinger Bridge framework for audio–visual deepfake localization that demonstrates the problems of symmetric fusion; IaMSB trades a smaller extractor and $\mathcal O(T)$ fusion module for finer temporal granularity to obtain more precise intervals. In future work, we will support adaptive budgets ($S_\mathrm{tgt}$), allowing the model to select the number of reasoning steps per instance. This enables differential resource allocation between authentic inputs and synchronous inputs.

\noindent
\textbf{Acknows:} This work is supported by the National Science and Technology Major Project of China (No. 2025ZD0219200), the National Natural Science Foundation of China (No. 62406225).

{
    \small
    \bibliographystyle{ieeenat_fullname}
    \bibliography{main}

@String(CVPR= {IEEE Conf. Comput. Vis. Pattern Recog.})

@String(ICCV= {Int. Conf. Comput. Vis.})

@String(ECCV= {Eur. Conf. Comput. Vis.})

@String(ICLR = {Int. Conf. Learn. Represent.})

@String(AAAI = {AAAI})

@String(CVPR  = {CVPR})

@String(ICCV  = {ICCV})

@String(ECCV  = {ECCV})

@String(TCSVT = {IEEE TCSVT})

@String(ICLR  = {ICLR})

@inproceedings{cai2022lavdf_dicta,
  title={Do you really mean that? content driven audio-visual deepfake dataset and multimodal method for temporal forgery localization},
  author={Cai, Zhixi and Stefanov, Kalin and Dhall, Abhinav and Hayat, Munawar},
  booktitle={International Conference on Digital Image Computing: Techniques and Applications (DICTA)},
  pages={1--10},
  year={2022}
}

@article{cai2023lavdf_cviu,
  author  = {Zhixi Cai and Shreya Ghosh and Abhinav Dhall and Tom Gedeon and Kalin Stefanov and Munawar Hayat},
  title   = {Glitch in the Matrix: A Large Scale Benchmark for Content Driven Audio--Visual Forgery Detection and Localization},
  journal = {Computer Vision and Image Understanding (CVIU)},
  volume  = {236},
  pages   = {103818},
  year    = {2023}
}

@inproceedings{zhang2023ummaformer,
  title={Ummaformer: A universal multimodal-adaptive transformer framework for temporal forgery localization},
  author={Zhang, Rui and Wang, Hongxia and Du, Mingshan and Liu, Hanqing and Zhou, Yang and Zeng, Qiang},
  booktitle={ACM International Conference on Multimedia (ACM MM)},
  pages={8749--8759},
  year={2023}
}

@article{liu2024avtfd,
  title={Audio-visual temporal forgery detection using embedding-level fusion and multi-dimensional contrastive loss},
  author={Liu, Miao and Wang, Jing and Qian, Xinyuan and Li, Haizhou},
  journal={IEEE Transactions on Circuits and Systems for Video Technology (TCSVT)},
  volume={34},
  number={8},
  pages={6937--6948},
  year={2023},
  publisher={IEEE}
}

@inproceedings{oorloff2024avff,
  title={Avff: Audio-visual feature fusion for video deepfake detection},
  author={Oorloff, Trevine and Koppisetti, Surya and Bonettini, Nicol{\`o} and Solanki, Divyaraj and Colman, Ben and Yacoob, Yaser and Shahriyari, Ali and Bharaj, Gaurav},
  booktitle={Proceedings of the IEEE/CVF Conference on Computer Vision and Pattern Recognition (CVPR)},
  pages={27102--27112},
  year={2024}
}

@inproceedings{gong2023cavmae,
  title={Contrastive Audio-Visual Masked Autoencoder},
  author={Gong, Yuan and Rouditchenko, Andrew and Liu, Alexander H and Harwath, David and Karlinsky, Leonid and Kuehne, Hilde and Glass, James R},
  booktitle = {Proceedings of the International Conference on Learning Representations (ICLR)},
  year      = {2023}
}

@inproceedings{huang2023mavil,
  author={Huang, Po-Yao and Sharma, Vasu and Xu, Hu and Ryali, Chaitanya and Li, Yanghao and Li, Shang-Wen and Ghosh, Gargi and Malik, Jitendra and Feichtenhofer, Christoph and others},
  title     = {MAViL: Masked Audio-Video Learners},
  booktitle = {Advances in Neural Information Processing Systems (NeurIPS)},
  volume={36},
  pages={20371--20393},
  year={2023}
}

@inproceedings{katamneni2024mmmsba,
  title={Contextual cross-modal attention for audio-visual deepfake detection and localization},
  author={Katamneni, Vinaya Sree and Rattani, Ajita},
  booktitle={IEEE International Joint Conference on Biometrics (IJCB)},
  pages={1--11},
  year={2024},
  organization={IEEE}
}

@article{koutlis2024dimodif,
  title={DiMoDif: Discourse modality-information differentiation for audio-visual deepfake detection and localization},
  author={Koutlis, Christos and Papadopoulos, Symeon},
  journal={arXiv preprint arXiv:2411.10193},
  year={2024}
}

@inproceedings{zhu2025regqav,
  title={Query-Based Audio-Visual Temporal Forgery Localization with Register-Enhanced Representation Learning},
  author={Zhu, Xiaodong and Wang, Suting and Yang, Junqi and Yang, Yuhong and Tu, Weiping and Wang, Zhongyuan},
  booktitle = {Proceedings of the ACM International Conference on Multimedia (ACM MM)},
  pages={8547--8556},
  year={2025}
}

@inproceedings{wu2024cfprf,
  title={Coarse-to-fine proposal refinement framework for audio temporal forgery detection and localization},
  author={Wu, Junyan and Lu, Wei and Luo, Xiangyang and Yang, Rui and Wang, Qian and Cao, Xiaochun},
  booktitle={ACM International Conference on Multimedia (ACM MM)},
  pages={7395--7403},
  year={2024}
}

@inproceedings{chugh2020mds,
  title={Not made for each other-audio-visual dissonance-based deepfake detection and localization},
  author={Chugh, Komal and Gupta, Parul and Dhall, Abhinav and Subramanian, Ramanathan},
  booktitle={ACM international conference on multimedia (ACM MM)},
  pages={439--447},
  year={2020}
}

@article{rahmath2025earlyexit,
  title={Early-exit deep neural network-a comprehensive survey},
  author={Rahmath P, Haseena and Srivastava, Vishal and Chaurasia, Kuldeep and Pacheco, Roberto G and Couto, Rodrigo S},
  journal={ACM Computing Surveys},
  volume={57},
  number={3},
  pages={1--37},
  year={2024}
}

@inproceedings{bajpai-hanawal-2024-ceebert,
  author    = {Bajpai, Divya Jyoti and Hanawal, Manjesh Kumar},
  title     = {CeeBERT: Cross-Domain Inference in Early Exit {BERT}},
  booktitle = {Findings of the Association for Computational Linguistics (ACL Findings)},
  pages     = {1736--1748},
  year      = {2024}
}

@inproceedings{Wang_2024_CVPR,
  author={Wang, Hongjie and Dedhia, Bhishma and Jha, Niraj K},
  title={Zero-TPrune: Zero-shot token pruning through leveraging of the attention graph in pre-trained transformers},
  booktitle = {Proceedings of the IEEE/CVF Conference on Computer Vision and Pattern Recognition (CVPR)},
  pages     = {16070--16079},
  year      = {2024}
}

@inproceedings{Xu_2024_WACV,
  title={Rethink cross-modal fusion in weakly-supervised audio-visual video parsing},
  author={Xu, Yating and Hu, Conghui and Lee, Gim Hee},
  booktitle={Proceedings of the IEEE/CVF Winter Conference on Applications of Computer Vision (WACV)},
  pages={5615--5624},
  year={2024}
}

@inproceedings{wei2024diagnosing,
  title={Diagnosing and Re-learning for Balanced Multimodal Learning},
  author={Wei, Yake and Li, Siwei and Feng, Ruoxuan and Hu, Di},
  booktitle={European Conference on Computer Vision (ECCV)},
  pages={71--86},
  year={2024}
}

@article{wu2024mlmm,
  title={Deep multimodal learning with missing modality: A survey},
  author={Wu, Renjie and Wang, Hu and Chen, Hsiang-Ting and Carneiro, Gustavo},
  journal={arXiv preprint arXiv:2409.07825},
  year={2024}
}

@inproceedings{yang2024facilitating,
  title={Facilitating multimodal classification via dynamically learning modality gap},
  author={Yang, Yang and Wan, Fengqiang and Jiang, Qing-Yuan and Xu, Yi},
  booktitle={Advances in Neural Information Processing Systems (NeurIPS)},
  volume={37},
  pages={62108--62122},
  year={2024}
}

@inproceedings{hu2025daf,
  title={Adaptive Multimodal Fusion: Dynamic Attention Allocation for Intent Recognition},
  author={Hu, Bo and Zhang, Kai and Zhang, Yanghai and Ye, Yuyang},
  booktitle={Proceedings of the AAAI Conference on Artificial Intelligence (AAAI)},
  volume={39},
  number={16},
  pages={17267--17275},
  year={2025}
}

@inproceedings{Cao_2024_CVPR,
  title={Madtp: Multimodal alignment-guided dynamic token pruning for accelerating vision-language transformer},
  author={Cao, Jianjian and Ye, Peng and Li, Shengze and Yu, Chong and Tang, Yansong and Lu, Jiwen and Chen, Tao},
  booktitle = {Proceedings of the IEEE/CVF Conference on Computer Vision and Pattern Recognition (CVPR)},
  pages={15710--15719},
  year      = {2024}
}

@inproceedings{avdeepfake1m2024,
  title={AV-Deepfake1M: A large-scale LLM-driven audio-visual deepfake dataset},
  author={Cai, Zhixi and Ghosh, Shreya and Adatia, Aman Pankaj and Hayat, Munawar and Dhall, Abhinav and Gedeon, Tom and Stefanov, Kalin},
  booktitle = {Proceedings of the ACM International Conference on Multimedia (ACM MM)},
  year      = {2024}
}

@inproceedings{araujo2025cavmaesync,
  title={CAV-MAE Sync: Improving Contrastive Audio-Visual Mask Autoencoders via Fine-Grained Alignment},
  author={Araujo, Edson and Rouditchenko, Andrew and Gong, Yuan and Bhati, Saurabhchand and Thomas, Samuel and Kingsbury, Brian and Karlinsky, Leonid and Feris, Rogerio and Glass, James R and Kuehne, Hilde},
  booktitle = {Proceedings of the IEEE/CVF Conference on Computer Vision and Pattern Recognition (CVPR)},
  pages={18794--18803},
  year      = {2025}
}

@inproceedings{ishikawa2025lgcavmae,
  title={Language-Guided Contrastive Audio-Visual Masked Autoencoder with Automatically Generated Audio-Visual-Text Triplets from Videos},
  author={Ishikawa, Yuchi and Nakada, Shota and Munakata, Hokuto and Saito, Kazuhiro and Komatsu, Tatsuya and Aoki, Yoshimitsu},
  booktitle={Proc. Interspeech},
  pages={2605--2609},
  year={2025}
}

@article{sun2024hicmae,
  title={Hicmae: Hierarchical contrastive masked autoencoder for self-supervised audio-visual emotion recognition},
  author={Sun, Licai and Lian, Zheng and Liu, Bin and Tao, Jianhua},
  journal={Information Fusion},
  volume={108},
  pages={102382},
  year={2024},
}

@article{shi2022avhubert,
  title={Learning audio-visual speech representation by masked multimodal cluster prediction},
  author={Shi, Bowen and Hsu, Wei-Ning and Lakhotia, Kushal and Mohamed, Abdelrahman},
  journal={arXiv preprint arXiv:2201.02184},
  year={2022}
}

@article{wang2024fas,
  title={Faster Diffusion Action Segmentation},
  author={Wang, Shuaibing and Wang, Shunli and Li, Mingcheng and Yang, Dingkang and Kuang, Haopeng and Qian, Ziyun and Zhang, Lihua},
  journal={arXiv preprint arXiv:2408.02024},
  year={2024}
}

@inproceedings{xu2024denoiseLoc,
  title={BOUNDARY DENOISING FOR VIDEO ACTIVITY LOCALIZATION},
  author={Xu, Mengmeng and Soldan, Mattia and Gao, Jialin and Liu, Shuming and P{\'e}rez-R{\'u}a, Juan Manuel and Ghanem, Bernard},
  booktitle = {Proceedings of the International Conference on Learning Representations (ICLR)},
  year      = {2024}
}

@inproceedings{hwang2025diffgebd,
  title={Generic Event Boundary Detection via Denoising Diffusion},
  author={Hwang, Jaejun and Gong, Dayoung and Kim, Manjin and Cho, Minsu},
  booktitle={Proceedings of the International Conference on Computer Vision (ICCV)},
  pages={14084--14094},
  year={2025}
}

@inproceedings{chen2025diffdvc,
  title={DiffDVC: Accurate Event Detection for Dense Video Captioning via Diffusion Models},
  author={Chen, Wei and Niu, Jianwei and Liu, Xuefeng and Wang, Zhendong and Tang, Shaojie and Zhu, Guogang},
  booktitle={Proceedings of the AAAI Conference on Artificial Intelligence (AAAI)},
  volume={39},
  number={2},
  pages={2221--2229},
  year={2025}
}

@inproceedings{liu2024lipslying,
  title={Lips are lying: Spotting the temporal inconsistency between audio and visual in lip-syncing deepfakes},
  author={Liu, Weifeng and She, Tianyi and Liu, Jiawei and Li, Boheng and Yao, Dongyu and Liang, Ziyou and Wang, Run},
  booktitle={Advances in Neural Information Processing Systems (NeurIPS)},
  volume={37},
  pages={91131--91155},
  year={2024}
}

@inproceedings{debortoli2021dsb,
  title={Diffusion schr{\"o}dinger bridge with applications to score-based generative modeling},
  author={De Bortoli, Valentin and Thornton, James and Heng, Jeremy and Doucet, Arnaud},
  booktitle={Advances in Neural Information Processing Systems (NeurIPS)},
  volume={34},
  pages={17695--17709},
  year={2021}
}

@inproceedings{deng2024reflectedSB,
  title={Reflected Schr{\"o}dinger Bridge for Constrained Generative Modeling},
  author={Deng, Wei and Chen, Yu and Yang, Nicole Tianjiao and Du, Hengrong and Feng, Qi and Chen, Ricky Tian Qi},
  booktitle={Uncertainty in Artificial Intelligence},
  pages={1055--1082},
  year={2024}
}

@article{mirsky2021deepfakes,
  title={The creation and detection of deepfakes: A survey},
  author={Mirsky, Yisroel and Lee, Wenke},
  journal={ACM computing surveys (CSUR)},
  volume={54},
  number={1},
  pages={1--41},
  year={2021}
}

@article{verdoliva2020media,
  title={Media forensics and deepfakes: an overview},
  author={Verdoliva, Luisa},
  journal={IEEE journal of selected topics in signal processing},
  volume={14},
  number={5},
  pages={910--932},
  year={2020}
}

@article{chesney2019deepfakes,
  title={Deep fakes: A looming challenge for privacy, democracy, and national security},
  author={Chesney, Bobby and Citron, Danielle},
  journal = {California Law Review (Cal. L. Rev.)},
  volume={107},
  pages   = {1753},
  year    = {2019}
}

@article{yang2023avoiddf,
  title={Avoid-df: Audio-visual joint learning for detecting deepfake},
  author={Yang, Wenyuan and Zhou, Xiaoyu and Chen, Zhikai and Guo, Bofei and Ba, Zhongjie and Xia, Zhihua and Cao, Xiaochun and Ren, Kui},
  journal={IEEE Transactions on Information Forensics and Security (TIFS)},
  volume={18},
  pages={2015--2029},
  year={2023}
}

@inproceedings{zhang2022actionformer,
  title={Actionformer: Localizing moments of actions with transformers},
  author={Zhang, Chen-Lin and Wu, Jianxin and Li, Yin},
  booktitle={European Conference on Computer Vision (ECCV)},
  pages={492--510},
  year={2022}
}

@inproceedings{adamw,
  title={Decoupled Weight Decay Regularization},
  author={Loshchilov, Ilya and Hutter, Frank},
  booktitle={International Conference on Learning Representations (ICLR)},
  year={2019},
}

@article{videomae,
  title={Videomae: Masked autoencoders are data-efficient learners for self-supervised video pre-training},
  author={Tong, Zhan and Song, Yibing and Wang, Jue and Wang, Limin},
  journal={Advances in neural information processing systems (NeurIPS)},
  volume={35},
  pages={10078--10093},
  year={2022}
}

@article{WavLM,
  title={Wavlm: Large-scale self-supervised pre-training for full stack speech processing},
  author={Chen, Sanyuan and Wang, Chengyi and Chen, Zhengyang and Wu, Yu and Liu, Shujie and Chen, Zhuo and Li, Jinyu and Kanda, Naoyuki and Yoshioka, Takuya and Xiao, Xiong and others},
  journal={IEEE Journal of Selected Topics in Signal Processing},
  volume={16},
  number={6},
  pages={1505--1518},
  year={2022},
  publisher={IEEE}
}
}


\end{document}